\documentclass[conference]{IEEEtran}
\usepackage{stmaryrd}
\usepackage{amsfonts}

\usepackage{setspace}
\usepackage{algorithm}
\usepackage{algorithmic}
\usepackage{boxedminipage} 
\usepackage{graphicx}
\usepackage{verbatim} 
\usepackage{amsmath}
\usepackage{times} 
\usepackage{gensymb}
\usepackage{dsfont}
\usepackage{multirow}

\usepackage[T1]{fontenc}
\usepackage{url}
\usepackage{amssymb}
\usepackage[tight,footnotesize]{subfigure}
\usepackage[caption=false,font=footnotesize]{subfig}
\usepackage{epsfig}

\IEEEoverridecommandlockouts    

\begin{document}

\title{\ \\ \LARGE\bf A Method For Dynamic Ensemble Selection Based on a Filter and an Adaptive Distance to Improve the Quality of the Regions of Competence\thanks{Rafael M. O Cruz, George D. C. Cavalcanti and Tsant Ing Ren are
with Center of Informatics, Federal University of Pernambuco, Recife, Brazil. (email: \{rmoc, gdcc,tir\}@cin.ufpe.br). Site: www.cin.ufpe.br/$\sim$viisar} \thanks{This work was supported in part by the Brazilian National
Research Council CNPq and by FACEPE}}

\author{Rafael M. O. Cruz, George D. C. Cavalcanti and Tsang Ing Ren}


\maketitle

\begin{abstract}

Dynamic classifier selection systems aim to select a group of classifiers that is most adequate for a specific 
query pattern. This is done by defining a region around the query pattern and analyzing the competence of the classifiers in this region. However, the regions are often surrounded by noise which can difficult the classifier selection. This fact makes the performance of most
dynamic selection systems no better than static selections. In this paper we demonstrate that the performance
of dynamic selection systems end up limited by the quality of the regions extracted. Thereafter, we propose a new dynamic classifier selection
system that improves the regions of competence in order to achieve higher recognition rates.
Results obtained from several classification databases show the proposed method not only 
significantly increase the recognition performance, but also decreases the computational cost.

\end{abstract}


\section{Introduction}

Multiple Classifier Systems/Ensemble of Classifiers have been widely 
studied in the past years as an alternative to increase efficiency and 
accuracy in pattern recognition problems~\cite{kittler,kuncheva}. The main motivation for using combination 
of classifiers derives from the observation that different classifiers usually 
commits errors in different patterns. The advantages of the individual (base) classifiers are combined
into a final solution. This leads to a system that presents more accurate results.
There are many examples in the literature that show the efficiency of ensemble
of classifiers in many tasks, such as, handwritten recognition~\cite{iwssip,ijcnn,bks},
signature verification~\cite{signverification} and image labeling~\cite{singh} .

There are two basic approaches for combination of multiple classifiers: selection 
and fusion. In the classifier fusion techniques~\cite{kittler,
quadratic,multioraveraging,fusionmatrix}, every classifier in the ensemble is used and the outputs are 
aggregated using a function (e.g. product rule, majority vote). Furthermore, another different 
classifier can be used to fuse the outputs~\cite{mixture,decisiontemplates,iwssip,ijcnn}.
In classifier selection the idea is to define a region of competence and search for the most competent 
or a subset with the most competent classifiers in the region. The selected classifier(s) is(are) used to give the final answer~\cite{lca,knora,classrank}. In some case it is possible to use a combination of selection and
fusion~\cite{selectionfusion}. These methods can also be static (the same combination for every pattern)
or dynamic (the combination depends on the query pattern).

\begin{figure}[htbp]
    \begin{center}
    	  \epsfig{file=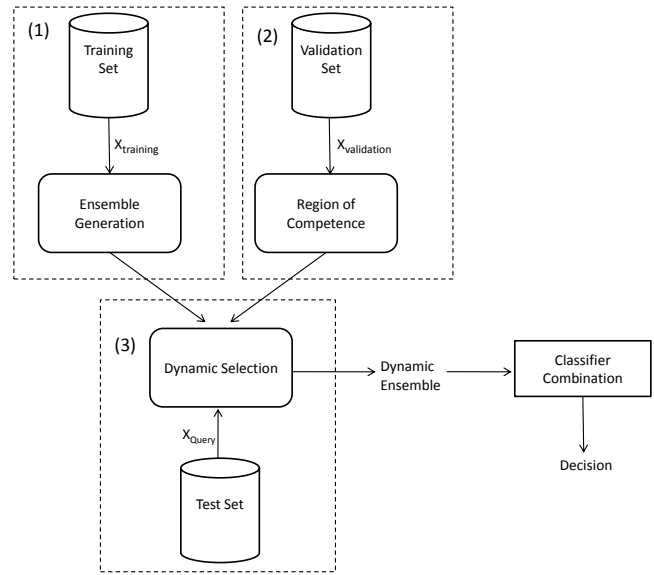, clip=,  width=0.48\textwidth}
    	  \end{center}
          \caption{Overview of dynamic classifier selection systems}
    \label{fig:dcss}
\end{figure}

The problem examined in this paper relies on dynamic classifier selection (DCS).
The classical dynamic classifier selection procedure is divided into
three levels~\cite{docs}: (1) Classifier generation which defines how the base classifiers are generated,
(2) Region of competence that is how to define the region in which the search for the best classifier is performed and
(3) Dynamic selection that defines the rule that selects the classifier(s), generated on the first level, based on the information extracted from the
regions defined on the second level. The classifier(s) selected on the third level is(are) used to classify the query pattern.
Figure~\ref{fig:dcss} shows an overview of a dynamic classifier selection system.

Many studies have been conducted on the first and on the third level. On the first level the most used algorithms are bagging~\cite{bagging}, boosting~\cite{boosting} and random subspaces~\cite{randomsub}. Duin~\cite{trainnottrain}
presented different ways to generate ensemble of classifiers and rank the methods according to its success. 

On the third level, Woods et al.~\cite{lca} proposed the DCS-LA (Dynamic Classifier Selection by Local Accuracy). 
In this technique, the accuracy of each classifier in the neighborhood of the test pattern is computed and the classifier
with the best result is selected. The classifier rank~\cite{classrank} approach is similar to the DCS-LA, but the selected 
classifier is the one that correctly classified more consecutive patterns in the neighborhood. 
Giancinto and Roli~\cite{mcb} proposed the Multiple Classifier Behavior algorithm which is a mixture 
of the DCS-LA with the behavior-knowledge space (BKS)~\cite{bks}. In this algorithm the local region
is measured based on the behavior of the classifiers. Kuncheva~\cite{clussel} used the overall local accuracy
on previously defined regions. During the test phase, the classifier with the highest accuracy in the desired
region is selected. 

However, given the fact that selecting only one classifier is very error prone, some researchers decided to
select a subset of classifiers. Ko et al.~\cite{knora} proposed an approach that 
aims to imitate the Oracle concept. The Oracle is
the upper limit of the ensemble performance~\cite{classfusion}. The KNORA-E (K Nearest ORAcles - Eliminate) which 
eliminates a classifier of the ensemble if the classifier misclassify any pattern of the neighbors. There is also a weighted version
KNORA-E-W that weights the outputs of the selected classifiers according to the distance between the query pattern and the neighbors. This work
also introduces two fusion algorithms: KNORA-U (K Nearest ORAcles - Union) and its weighted version KNORA-U-W.
Soares et al.~\cite{SoaresSCS06} select the $N$ most accurate classifier, based on a defined region of competence,
and the $J$ most diverse classifiers to create the ensemble. The values of $N$ and $J$ were defined by the authors.
These techniques are called dynamic ensemble selection (DES) as they can select more than one classifier.

However, not much attention have been given to the second level (region of competence) in how the quality of this
region influences the final result. The rule defined for selecting the classifiers (third level) depends on 
the quality of the information obtained from the region of competence. The dynamic selection should probably fail 
if there are many noisy patterns in the region of competence. 

The focus of this paper is on the second level. First we show the performance of dynamic classifier selection 
is limited by the quality of the region of competence (how it is defined). A practical example is used to illustrate cases when the
dynamic classifier/ensemble selection systems fail because of noises in the region of competence. Also we compare the recognition
performance of the techniques with the algorithm that defines the region of competence and show
that the results are really close. In some cases the dynamic classifier selection results are even slightly inferior. 

Based on this analysis, we propose a new dynamic ensemble selection technique that achieve more accurate results by improving the quality of the regions of competence. This is performed using two strategies:
One is a filter that removes samples that are considered noise, creating soft decision boundaries. The other is an
adaptive version of the k-Nearest Neighbor algorithm that uses weights to indicate whether a pattern is close
to patterns of different classes or not. The objective is to eliminate noisy patterns before the execution
of the classifier selection (third level). Thus, improving the overall system performance. 

In order to demonstrate the efficiency of the proposed approach, we conducted experiments using nine classification problems. We show that the performance of previous techniques becomes limited by the performance of the algorithm that creates the region of competence. Thereafter, we show the proposed technique not only increases the recognition rate but also can decrease the computational time as it becomes easier for the system to select the best classifiers.

This paper is organized as follow. An analysis of how the regions of competence influences the classifier selection is shown in Section~\ref{algo}. Section~\ref{neigh} describes the proposed system. The experiments are presented in Section~\ref{exp} and the conclusion is shown in Section~\ref{conc}.

\section{Analysis of the Influence of the Region of Competence}
\label{algo}

The influence of the region of competence in DCS system is analyzed in this section. In order to do so, first
we explain the KNORA-ELIMINATE algorithm~\cite{knora}. The KNORA-E was selected because it performs slight better than
the other dynamic selection schemes~\cite{knora}.
Thereafter, we perform an analysis of the influence of the quality of the region of competence using a practical example.

\subsection{KNORA-ELIMINATE}

This approach explores the oracle concept to dynamically select the classifiers.
Let $X_{i}, \: i = 1, \cdots, \: k$ be the $k$ nearest neighbors of the query pattern $X$ and an ensemble
of $L$ classifiers $C_{j}, \: j = 1, \cdots, \: L$, the dynamic ensemble $E^{*}$ is composed of
the classifiers $C_{j}$ that correctly classifies every neighbor $X_{i}$. Classifiers that misclassify
any of the $k$ neighbors are eliminated. If none classifier can correctly classify every neighbor,
the value of $k$ is decreased and the rule continues the search until at least one classifier correctly
classifies all the neighbors. 

One advantage of this method is that the number of neighbors is not fixed, although it can only decrease. However, the cost
of reducing the neighborhood and recalculate the method is computationally expensive. Like the other 
dynamic techniques, this rule is very dependent on the quality of the neighborhood.

\subsection{Analysis}

\begin{figure*}[ht]
	  \centering
	  	 
	  	 \subfigure[Banana dataset]{\includegraphics[width=2.8in]{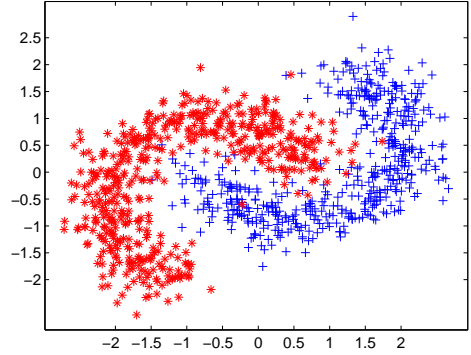}} \\
	     \subfigure[Validation Set]{\includegraphics[width=2.8in]{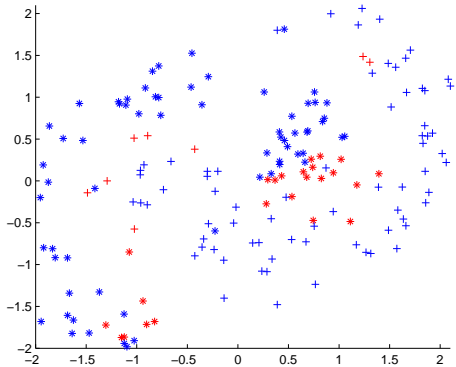}} \quad
	     \subfigure[Errors]{\includegraphics[width=2.8in]{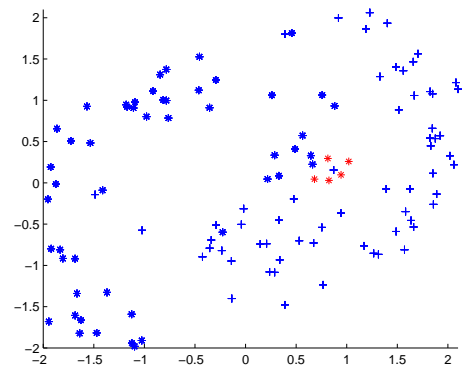}}  
	   \caption{Problems with the neighborhood information}
	   
 \label{fig:neighborhood}
\end{figure*}

To demonstrate the problem that the dynamic classifier/ensemble selection techniques have with the quality of the region of competence, we performed an experiment using an ensemble of 10 Perceptrons generated using the bagging algorithm. A neighborhood of $k = 7$ is used. 
Figure~\ref{fig:neighborhood} shows the misclassifications obtained by the KNORA-E for the
Banana dataset. Figure~\ref{fig:neighborhood}(a) shows the form of the Banana dataset. Figure~\ref{fig:neighborhood}(b) shows the
errors obtained in this dataset (in red) and the validation set (in blue). The validation dataset is used to compute
the region of competence. Figure~\ref{fig:neighborhood}(c) shows some patterns of the class \textbf{$\ast$} (in red) that although they are closer to its class mean, they were misclassified because there is a pattern from the other class \textbf{$+$} among them. This pattern is closer
to the other class mean \textbf{$\ast$} than its own class mean \textbf{$+$}. Thus it can be considered a noise.

The current dynamic ensemble selection systems fail when situations like this happens. The current systems end up selecting the wrong classifiers when there are noisy patterns near the query pattern as the classifier that can recognize those noise patterns and therefore achieve the highest accuracy in the
neighborhood probably have overfitted in the region. That explain why the dynamic selection methods become limited to the performance of the algorithm that defines the region of competence. Thus, if we improve the quality of the neighborhood, the performance of the dynamic classifier/ensemble selection method will also improve. This is an important point in the recognition rate of the system that did not receive much 
attention. In the experiments section we demonstrate the limitation imposed by the performance of the algorithm that defines the region of competence using empirical results. 

\section{The Proposed Approach: DES-FA}
\label{neigh}

\begin{figure*}[htbp]
    \begin{center}
    	  \epsfig{file=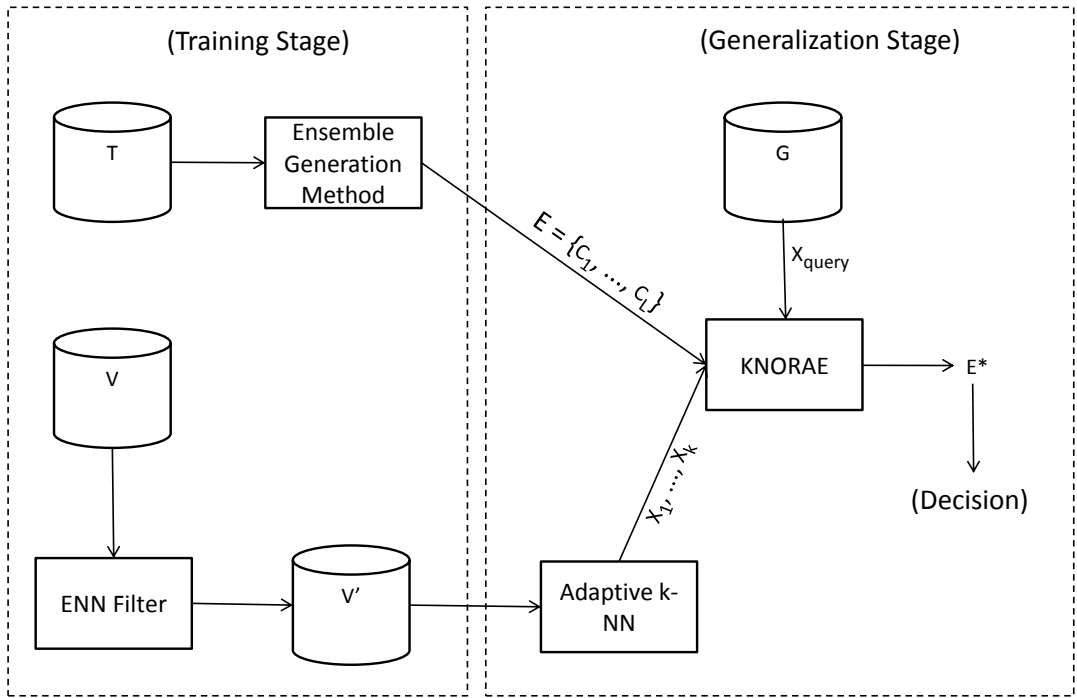, clip=, width=0.80\textwidth}
    	  \end{center}
          \caption{Overview of the DES-FA system.}
    \label{fig:blocdiagram}
\end{figure*}

In this section the proposed ideas to improve the quality of the neighborhood and consequently the dynamic selection are shown. 
Two techniques were used. First, a noise reduction filter is applied to the validation dataset (dataset where the regions of competence are computed) to remove noisy patterns. This step is done during the training procedure. Thereafter, a variation of the k-Nearest Neighbor algorithm is proposed in order to improve the quality of the computed neighbors. Figure~\ref{fig:blocdiagram} shows an overview of the proposed system. $T$ is the training set,
$V$ the validation dataset and $G$ the test dataset (generalization). During the training stage, the ensemble $E = \left \{ C_{1}, \, \cdots C_{L} \right \}$ is generated using the dataset $T$. The Edited Nearest Neighbor (ENN) filter~\cite{enn} is applied to the validation dataset $V$ generating the dataset $V', \left | V' \right |  \leq \left |  V  \right |$. The ENN filter works eliminating noise on the decision boundaries. Thus the algorithm produces soft decision boundaries. 

In the test phase, the local region is computed using the adaptive k-NN algorithm~\cite{aknn} using the patterns of the filtered dataset $V'$. The adaptive k-NN is a variation of the traditional k-NN that uses weights to indicate how close a training pattern is from patterns of different classes. The weight is used in order to have a higher probability of selecting patterns that are distant from the border. Thus, patterns with higher probability of being noise are less likely to be chosen. On the third stage (classifier selection) we use the KNORA-Eliminate rule~\cite{knora} to select the dynamic ensemble $E^{*}$ using the region of competence defined by the adaptive k-NN algorithm. We call the proposed system DES-FA (Dynamic Ensemble Selection by Filter + Adaptive Distance). The ENN filter and the Adaptive k-NN are described in the next sections.

\subsection{Edited Nearest Neighbor Filter}

\begin{figure}[htbp]
    \begin{center}
    	  \epsfig{file=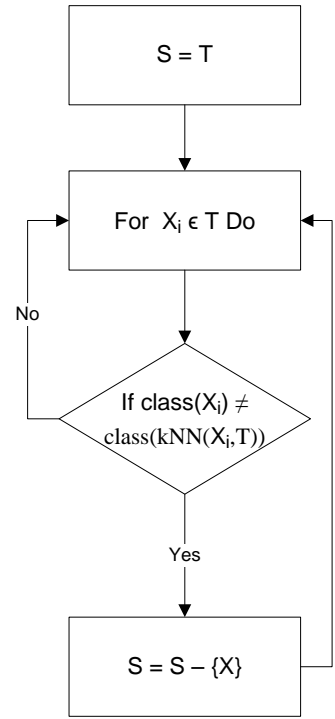, clip=,  height=0.40\textheight}
    	  \end{center}
          \caption{FlowChart of the ENN algorithm}
    \label{fig:flowchartenn}
\end{figure}

The edited nearest neighbor rule~\cite{enn} works as a noise reduction filter to create 
smoother class boundaries. The central points of the classes are preserved. Figure~\ref{fig:flowchartenn} and
Algorithm~\ref{alg:enn} show the steps of the ENN algorithm. The algorithm works as follow: Let $T$ be the training set, and
$S$ the filtered set, the algorithm perform the nearest neighbor classification for each $X_{i} \in T$ 
using $T$ as reference. If $X_{i}$ is misclassified using the k-NN algorithm, it is considered a noise and removed from the
final set $S$.

\begin{algorithm}[h]
	\caption{The Edited Nearest Neighbor Algorithm}
	\label{alg:enn}
		\begin{algorithmic}[1]
		\REQUIRE Training Set $T$
		\STATE $S = T$
		\FOR {each $X_{i} \in T$}
		\IF {$ class\left ( X_{i} \right ) \neq class\left (kNN \left (  X_{i},T \right ) \right ) $ }
			\STATE $S = S - \left \{  X_{i} \right \}$
		\ENDIF
		\ENDFOR
		\RETURN $S$
		\end{algorithmic}
\end{algorithm}

\begin{figure*}[ht]
	  \centering
	  	 \subfigure[Original]{\includegraphics[width=2.8in]{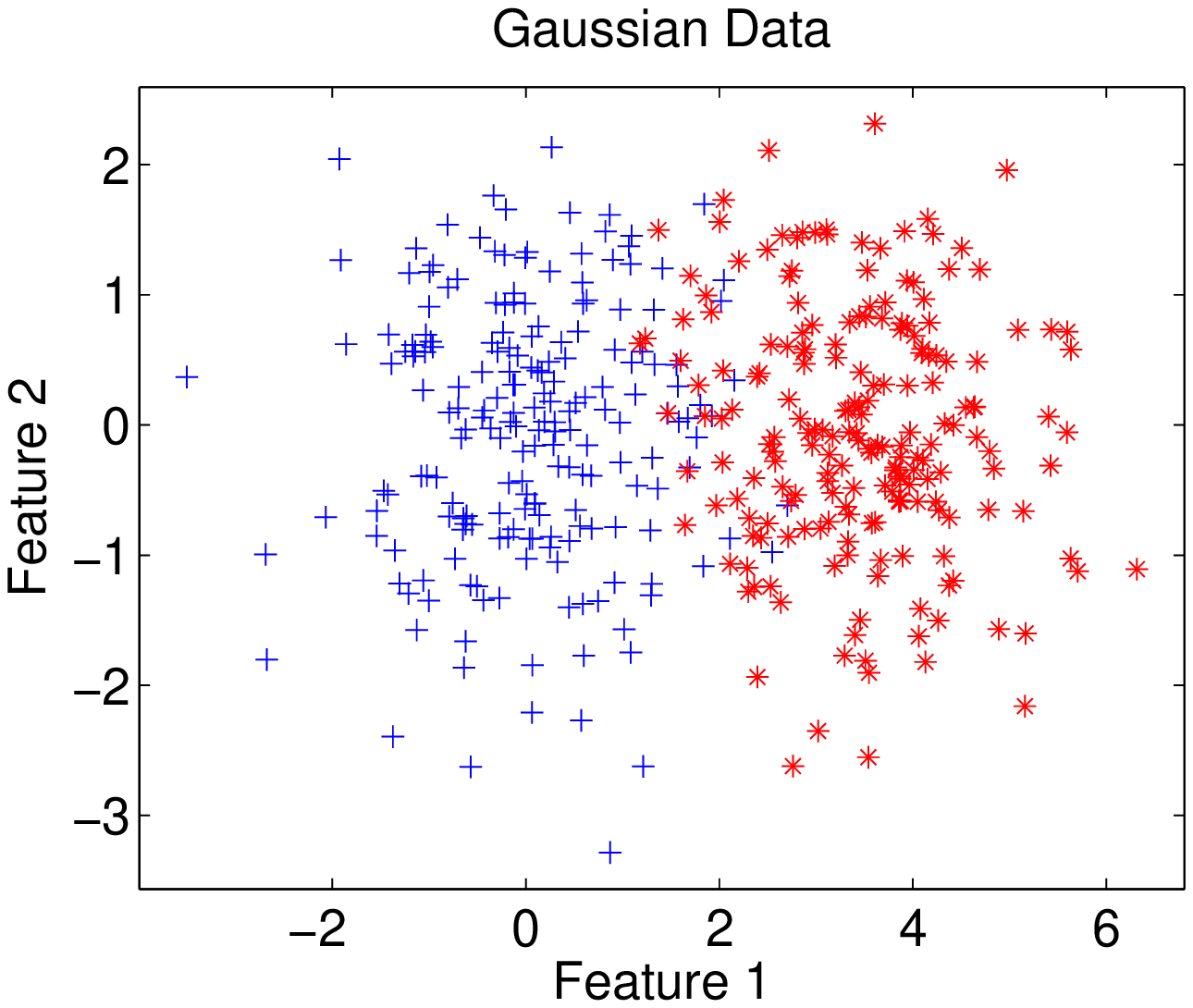}}    \qquad 
	     \subfigure[ENN k = 1]{\includegraphics[width=2.8in]{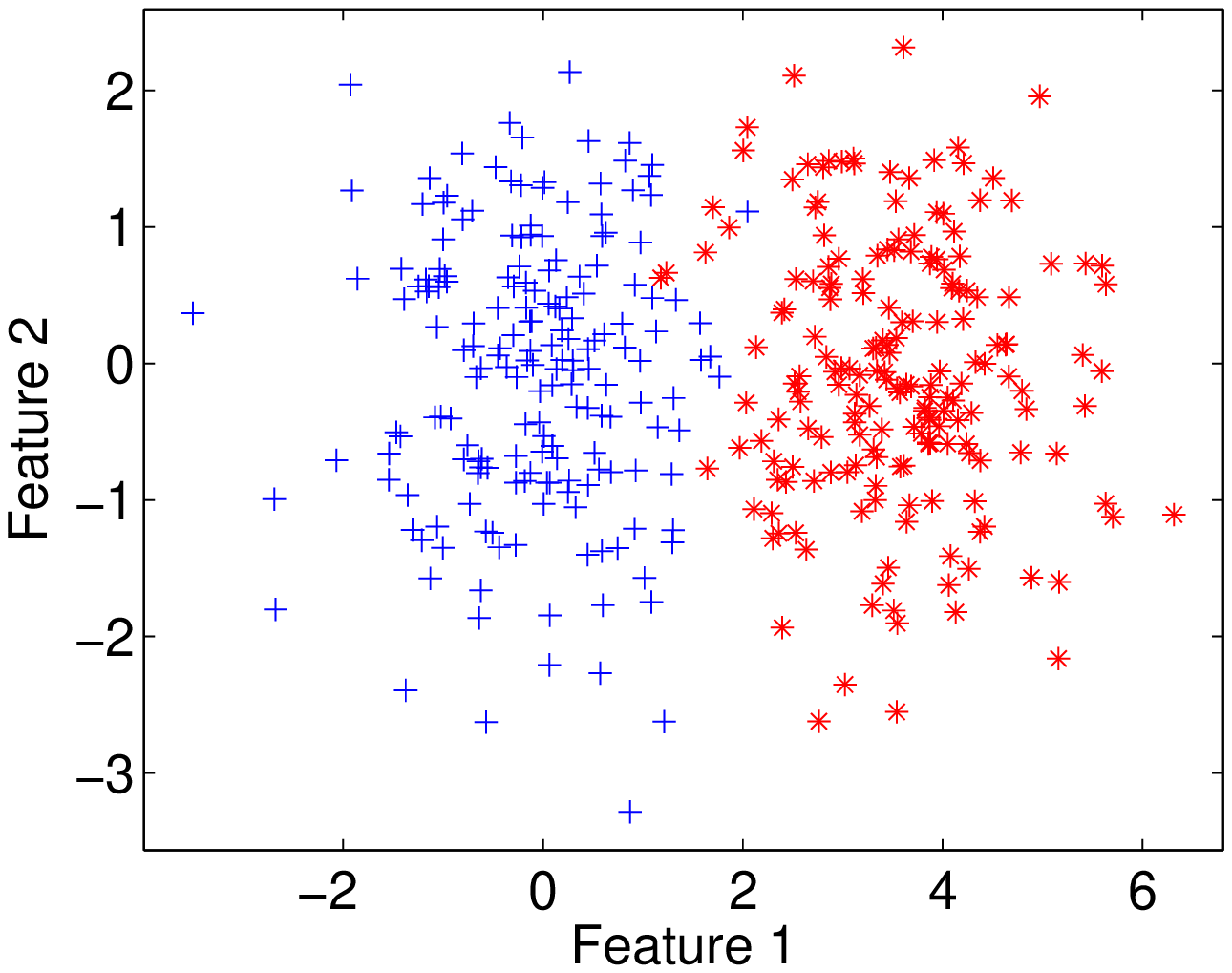}}  \\
	     \subfigure[ENN k = 3]{\includegraphics[width=2.8in]{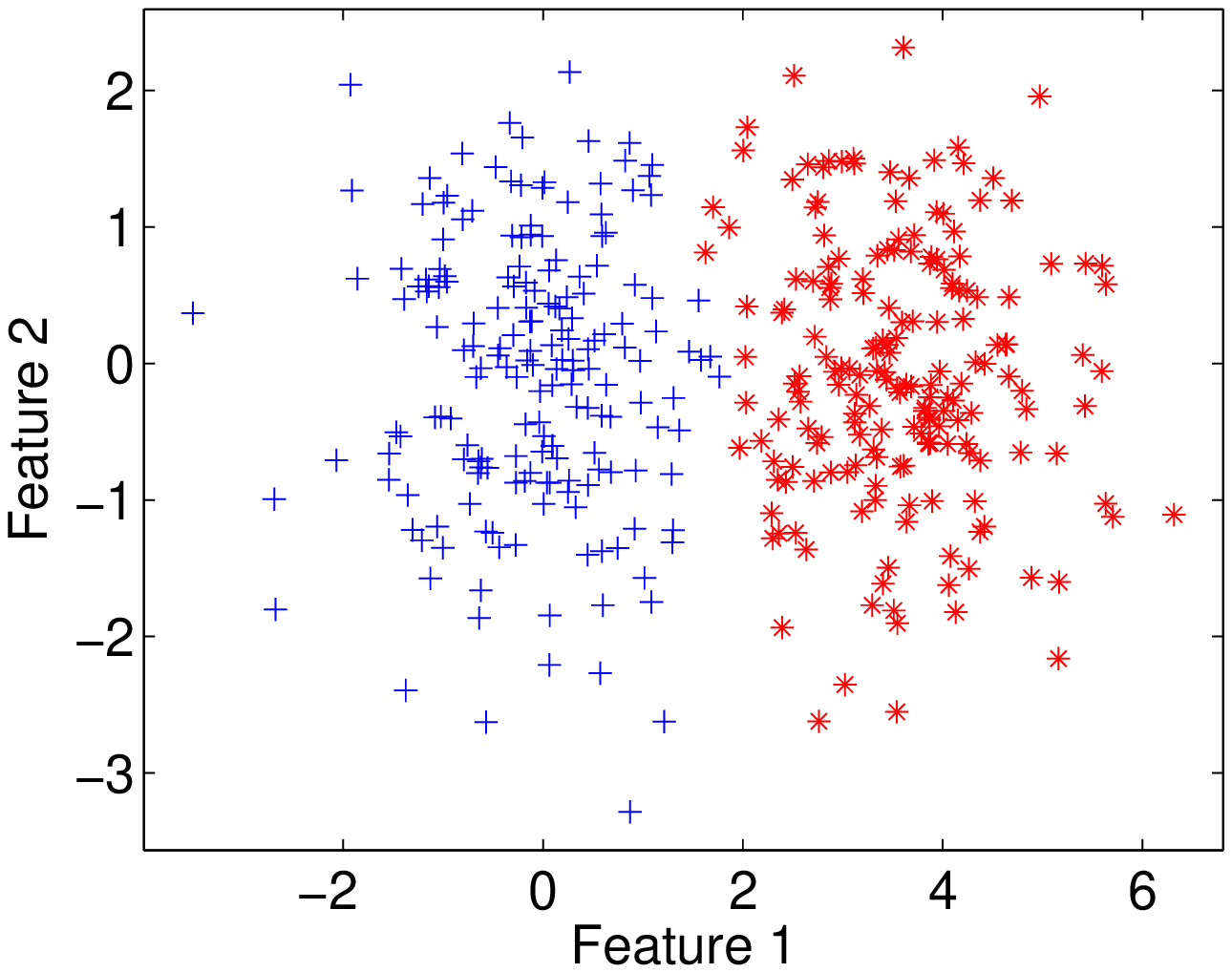}}   \qquad 
	     \subfigure[ENN k = 5]{\includegraphics[width=2.8in]{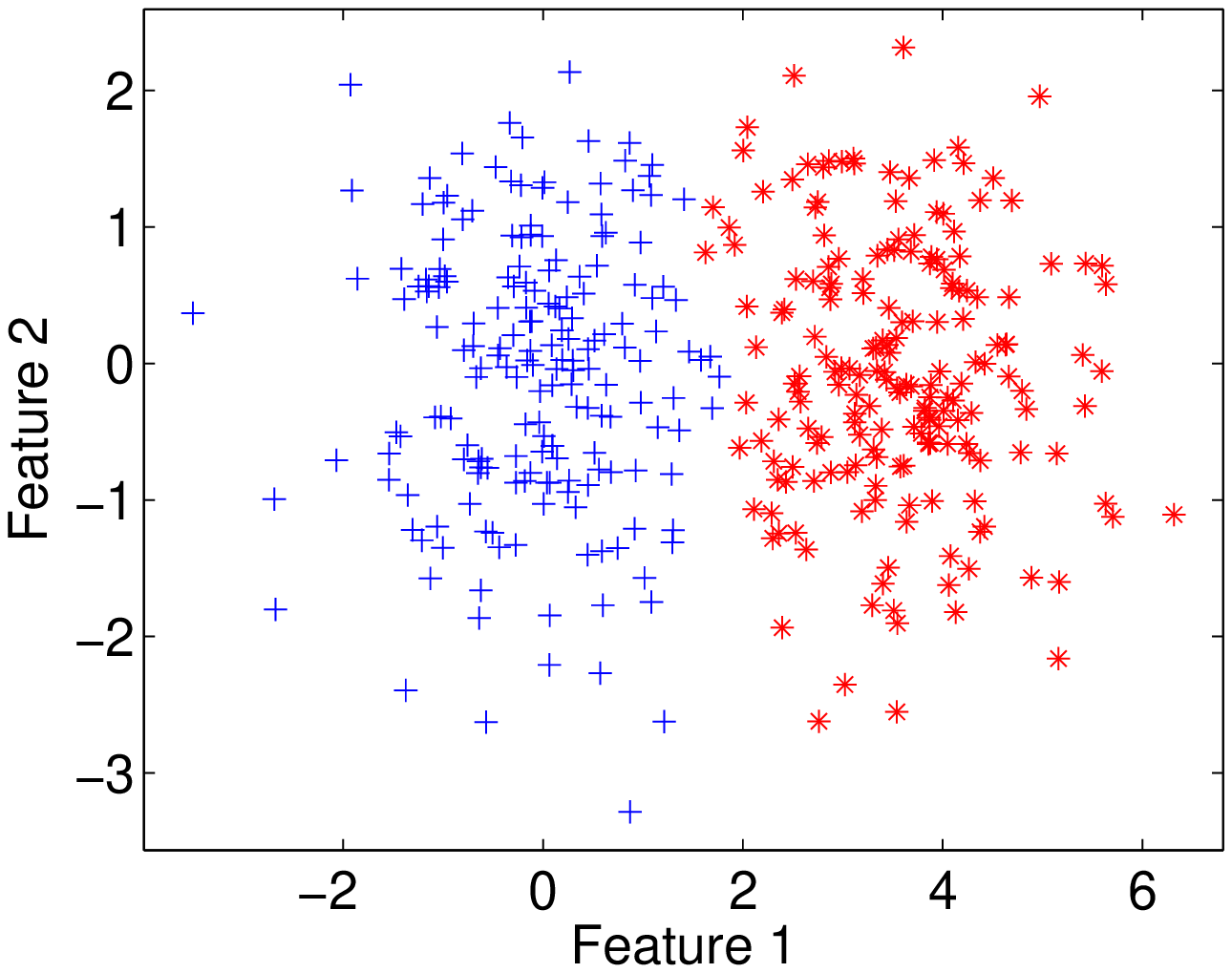}}
	   \caption{Results of the ENN algorithm for two gaussian distributions}
 \label{fig:enn}
\end{figure*}

Figure~\ref{fig:enn} shows an example of the application of the ENN filter. The data was constructed using two Gaussian distributions generated with
$\mu_{1} = \left [ 0.0,\,   0.0 \right ]$, $\mu_{2} = \left [ 3.5,\, 0.0 \right ]$ and
$\sigma_{1}^{2} = \sigma_{2}^{2} = 1$. Figure~\ref{fig:enn}(a) shows the original distribution.
Figures~\ref{fig:enn}(b), (c) and (d) present the result after the execution of the ENN algorithm
with $k =1, \, 3$ and $5$ respectively.

\subsection{K-Nearest Neighbor with Adaptive Distance}
\label{adapt}

The adaptive distance~\cite{aknn} calculates, for each 
training sample $X_{i}$, the largest sphere centered on $X_{i}, \: i = 1, \cdots, \: N$ 
that excludes every training pattern of different classes $X_{j}, \: j = 1, \cdots, \: N$. This is performed by 
computing the minimum distance (sphere radius) $R_{i}$ between the training pattern $X_{i}$ 
and the training samples of different classes (Equation~\ref{eq:radius}).  
With the radius $R_{i}$, the adaptive distance  between the test pattern $X_{test}$ 
and $X_{i}$ is defined by equation~\ref{eq:adapDist}. The distance $d(X_{test}, \: X_{i})$
can be any distance, such as, the Euclidean or the Manhattan distance.

\begin{equation}
\label{eq:radius}
R_{i} = \min d\left ( X_{i}, X_{j} \right ),  c_{i} \neq c_{j}
\end{equation}

\begin{equation}
\label{eq:adapDist}
D_{adap}(X_{test},X_{i}) = \frac{d(X_{test},X_{i})}{R_{i}}
\end{equation}

Using this method, samples closer to its class mean 
have bigger radius ($R_{i}$) than samples that are near the class boundaries. Thus, samples that
are closer to the class boundaries become more distant to the query pattern while the 
ones next to the class means becomes closer. Therefore, the probability of selecting a
noise as neighbors is lower.

The idea behind using the ENN filter and the adaptive k-NN techniques comes from the fact that they reduce the number of undesirable patterns
in the region of competence. However, it is not guaranteed that the ENN will eliminate every undesirable pattern.
The adaptive k-NN works in a way that pattern closer to the decision boundaries and therefore more probable of being
noise have less chance of being selected. Therefore even if an undesirable pattern was not eliminated using the ENN, the probability of selecting this pattern using the adaptive k-NN is lower. Thus, it is interesting to use both techniques as one can overcome the limitation of the other.




\section{Experiments}
\label{exp}

To ensure the efficiency of the proposed DES-FA, the experiments were conducted using nine databases, 
seven from the UCI machine learning repository\footnote[1]{http://archive.ics.uci.edu/ml} 
and two artificially generated using the Matlab PRTOOLS toolbox\footnote[2]{www.prtools.org}. The key 
features of the databases are shown in Table~\ref{table:datasets}. 
The ensemble is generated using the bagging technique which is described below.

\begin{table}[htbp]
    \centering
    \caption{Features of the selected databases}
     \label{table:datasets} 
     \resizebox{0.48\textwidth}{!}{
     \begin{tabular}{|c|c|c|c|}
    \hline
      \textbf{Database} & \textbf{ No. of Instances} & \textbf{Dimensionality} & \textbf{No. of Classes} \\
        \hline

        \textbf{Pima} & 768 &	8 &	2 \\ \hline

        \textbf{Liver Disorders} & 345 &	6 &	2  \\ \hline

        \textbf{WDBC} & 568 &	30 &	2 \\ \hline
      
        \textbf{OptDigits} & 5620 & 64 & 10 \\ \hline
        
        \textbf{Blood transfusion} & 748 &	4 &	2  \\ \hline
       
        \textbf{Image Segmentation} & 2310  & 19 & 7 \\ \hline
  
        \textbf{Banana} & 600 &	2 &	2 \\ \hline
        
        \textbf{Vehicle} & 846 &	18 &	4 \\ \hline
   
        \textbf{Lithuanian classes} & 600 &	2 &	2  \\ 

    \hline
    \end{tabular}
    }
\end{table}

\subsection{Bagging}

Bagging is an acronym for Bootstrap AGGregatING~\cite{bagging}. The idea behind bagging is 
to simply build a diverse set of classifiers by selecting different subsets of the training set to train 
the base classifiers. The subsets are generated randomly. The diversity 
among the classifiers is achieved by the use of different training sets. One important point 
using this technique is the fact that the base classifiers should be 
unstable. A classifier is considered unstable if small perturbations in the training set results
in large changes in the constructed predictor~\cite{breiman96bias}. In general, classifiers that presents
high variance such as Neural Networks and Decision Trees are unstable. Linear Discriminant and
k-Nearest neighbor are considered stable classifiers. Also it is known that bagging 
presents good results when used with weak classifiers~\cite{baggingfor}. Algorithm~\ref{alg:bagging} summarizes the steps of the bagging algorithm.

\begin{algorithm}[h]
	\caption{The Bagging Algorithm}
	\label{alg:bagging}
		\begin{algorithmic}[1]
		\REQUIRE Training Set $T$
		\FOR {$i = 1$ \TO $L$}
		
			\STATE Take a bootstrap $T*$ from $T$
			\STATE Train $C_{i}$ with $T*$
			\STATE $E = E \cup C_{i}$
			
		\ENDFOR
		\RETURN $E$	
		\end{algorithmic}
\end{algorithm}

\subsection{Results}

A total of 20 iterations using different divisions between training/test were used for each dataset.
The datasets were divided into $50\%$ for the training set and $50\%$ for the test set. The only exceptions were
the Optical Digits and the Image segmentation datasets as the training and test set have been defined by the UCI repository. 
The training set was divided into $75\%$ for training and $25\%$ for validation. The validation dataset is used
to compute the regions of competence.
The ensemble is composed of 10 Perceptron and the number of neighbors is empirically 
set to 7. The Perceptron classifier was selected because it is unstable and a weak model. Therefore it is
suitable to be used with Bagging. Majority Vote rule~\cite{kittler} was selected as combination rule.

First, we show a comparison of the KNORA-E with the performance of the k-NN algorithm using the leave-one-out methodology~\cite{Stone74} to demonstrate that the current dynamic classifier selection systems are limited by the performance of the region of competence algorithm.
This methodology uses only one pattern as the test and the remaining as the training data. This is repeated until every pattern of
training data is used as test. Thus, using this test, we have the percentage of patterns that have a "bad" neighborhood. They are the patterns that
the dynamic selection techniques mostly presents error.

The comparison is shown in Table~\ref{table:leaveoneout}. It can be observed that the performance of the KNORA-E
is close to the results of the leave-one-out on the datasets and in some cases, the results are even slightly inferior. The result of the
KNORA-E is even lower than the best classifier of the ensemble (Single Best) or the static ensemble for some datasets. This demonstrates that the dynamic selection methods are being limited to the performance of the local region algorithm. The classifier selection does not produces accurate results if the results extracted from the region of competence is not accurate enough. These results show how important
the definition of region of competence is and the importance of putting efforts in the design of a better region of competence.

\begin{table}[htbp]
    \centering
    \caption{Comparison between the Dynamic Ensemble Selector and the Leave-One-Out result.}
     \label{table:leaveoneout} 
     \resizebox{0.48\textwidth}{!}{
     \begin{tabular}{|c|c|c|c|c|}
    \hline
      \textbf{Database} & \textbf{Leave-One-Out} & \textbf{KNORA-E} &  \textbf{Static Ensemble}  & \textbf{Oracle}  \\
        \hline

        \textbf{Pima} & 73.05 & 73.16 &  73.28 & 95.10 \\ \hline

        \textbf{Liver Disorders} & 65.80  & 63.86 & 62.76 & 90.07 \\ \hline
        
        \textbf{WDBC} & 97.02 & 96.93 & 96.35 & 99.13 \\ \hline
        
        \textbf{Optical Digits} & 85.97 & 79.32 & 81.47 & 91.84 \\ \hline
        
        \textbf{Image Segmentation} & 85.72 & 59.09 & 65.27 & 89.97 \\ \hline
        
        \textbf{Banana} & 90.27 & 88.83 & 81.43 & 94.75 \\ \hline
        
        \textbf{Vehicle} & 72.35 & 81.19 & 82.18 & 96.80 \\ \hline
        
        \textbf{Lithuanian Classes} & 91.02 & 88.83 & 82.33 & 98.35 \\ \hline
  
        \textbf{Blood transfusion} & 74.74 & 74.59 & 75.24 & 94.20  \\ 

    	\hline
    \end{tabular}
    }
\end{table}   

\begin{table*}[htbp]
    \centering
    \caption{Comparative results using Perceptron as weak classifier. The results are the mean and the standard deviation obtained over 20 iterations}
     \label{table:ResultsENNAKK} 
     \resizebox{6.6in}{!}{
     \begin{tabular}{|c|c|c|c|c|c|c|c|}
    \hline
      \textbf{Database} & \textbf{DES-FA (1)} & \textbf{DES-FA (3)} & \textbf{DES-FA (5)} & \textbf{A-k-NN + KNORA-E} & \textbf{KNORA-E} & \textbf{Static Ensemble} &  \textbf{Oracle}\\
        \hline

        \textbf{Pima} & 74.89(1.63) &	75.35(1.37) &	\textbf{76.04(1.61)} &	74.02(1.57) &	73.16(1.86) & 73.28(2.08) & 95.10(1.19) \\ \hline

        \textbf{Liver Disorders} & \textbf{65.72(3.81)} &	65.49(3.39) &	65.23(4.07) &	63.98(3.418) &	63.86(3.284) & 62.76(4.81) & 90.07(2.41) \\ \hline

        \textbf{WDBC} & 96.77(1.11) &	96.40(0.95) &	96.46(1.13) &	\textbf{97.18(1.13)} &	96.93(1.10) & 96.35(1.14) & 99.13(0.52) \\ \hline

        \textbf{Optical Digits} & 83.65(2.63) &	84.73(3.51) &	82.84(3.40) &	\textbf{86.78(3.20)} &	79.32(3.47) & 81.47(4.67) & 91.84(2.03)\\ \hline

        \textbf{Blood Transfusion} & \textbf{77.35(0.97)} &	76.17(1.56) &	76.42(1.16) &	75.21(2.10) &	74.59(2.62) & 75.24(1.67) & 94.20(2.08) \\ \hline

        \textbf{Image Segmentation} & \textbf{88.74(0.70)} &	63.88(6.62) &	80.45(3.25) &	66.16(5.47) &	59.09(11.32) & 65.27(3.32) & 89.97(3.46)  \\ \hline
 
        \textbf{Banana} & \textbf{90.16(3.18)} &	89.16 2.25) &	89.57(2.65) &	89.93(2.87) &	88.83(1.67) & 81.43(3.92) & 94.75(2.09) \\ \hline

        \textbf{Vehicle} & 71.7(4.11) &	80.00(2.21) &	80.20(4.05) &	80.29(1.45) &	\textbf{81.19(1.54)} & 82.18(1.31) & 96.80(0.94) \\ \hline

        \textbf{Lithuanian Classes} & 92.16(2.61) &	\textbf{92.23(2.46)} &	91.65(2.37) &	92.16(2.73) &	88.83(2.50) & 82.33(4.81) & 98.35 (0.57) \\

    \hline
    \end{tabular}
    }
    
\end{table*}

The comparison of the KNORA-E with the proposed DES-FA is shown in Table~\ref{table:ResultsENNAKK}. The number inside parenthesis is the value of the parameter $k$ used in the ENN filter ($k =1, \, 3$ and $5$). 
The KNORA-E is compared with the version using only the adaptive k-NN and with the DES-FA. The ENN was evaluated with
$k =1, \, 3$ and $5$. The performance of the single best classifier, static ensemble and the Oracle are also shown for comparison.

Only one out of nine datasets the KNORA-E algorithm presented the best result (Vehicle dataset). In most
cases the DES-FA presented the best results. It is also important to observe
that the the adaptive k-NN improved the result upon the standard algorithm in eight datasets. For the Optical digits
and the WDBC the adaptive k-NN alone presented better results than the DES-FA (although the DES-FA still improves upon the
KNORA-E). The ENN filter probably removed some important patterns in these datasets. A paired t-test with 95\% confidence was performed to better
compare the performance of the methods. The results of the DES-FA over the Pima, Liver Disorders, Image Segmentation, Banana and Lithuanian datasets showed statistically better than the KNORA-E technique. For the other datasets the difference between the DES-FA and the KNORA-E is not statistically different. However, the mean accuracy obtained by the DES-FA is higher.

It is important to mention that one of the problems of the KNORA-E algorithm is the computational cost of reducing the neighborhood. When
none of the base classifiers correctly classifies all the neighbors, the neighborhood is reduced and the algorithm 
computes again. This becomes a problem when there are many noisy patterns in the dataset. The algorithm needs to reduce the neighborhood
often, which increases considerably the computational time. Using the ENN filter and the adaptive k-NN, less noise
are selected as neighbors (some are eliminated by the ENN and some are not selected by the adaptive k-NN rule). Therefore the number of times that
the KNORA-E algorithm needs to reduce the neighborhood decreases considerably. Also the ENN rule eliminates some patterns of the
validation set which contributes in decreasing the cost of computing the nearest neighbor. Table~\ref{table:ProcessingTimeENN} shows the
processing time (time to process the whole database) obtained by the KNORA-E algorithm and the DES-FA.   
In most datasets the processing time is much lower and it is explained by the better quality of the selected region of competence.

\begin{table}[htbp]
    \centering
    \caption{Average Processing Time (Seconds)}
    \label{table:ProcessingTimeENN} 
    \resizebox{2.5in}{!}{
    \begin{tabular}{|c|c|c|}
    \hline
      \textbf{Database} & \textbf{DES-FA (k)} & \textbf{KNORA-E}  \\ 
        \hline

        \textbf{Pima} & \textbf{91.71} &	177.00 \\ \hline

        \textbf{Liver}& \textbf{75.82} &	103.52 \\ \hline

        \textbf{Breast}  & \textbf{48.94} &	64.66 \\ \hline

        \textbf{Optical Digits}  & \textbf{599.09} &	1609.00 \\ \hline

        \textbf{Blood Transfusion} & \textbf{78.59} &	222.30 \\ \hline

        \textbf{Segmentation} & 696.83 &	\textbf{288.03} \\ \hline
 
        \textbf{Banana} & \textbf{58.95} &	100.59 \\ \hline

        \textbf{Vehicle}  & \textbf{122.03} &	150.97 \\ \hline

        \textbf{Lithuanian Classes} & \textbf{61.60} &	82.55 \\
    \hline
    \end{tabular}
    }
\end{table}

\section{Conclusion}
\label{conc}

In this paper the problem of dynamic classifier selection is discussed. The paper is focused in how the regions of competence
influences the performance of the system and two strategies are proposed in order to achieve better results.
We demonstrate that the performance of the ensemble selection methods is very dependent to the performance of the algorithm that
defines the regions of competence. Based on that two techniques for improving this information is shown. One that works
as a filter eliminating undesirable patterns and the other is a variation of the nearest neighbor algorithm that turns patterns
that are more probable of being noise more difficult to be selected. These techniques are used together in order to enhance the quality of the extracted region of competence. 

Experiments were conducted over nine different datasets. Results show the proposed technique improves the
recognition rates for eight of the nine datasets.
We believe this idea can be used to improve recognition rates for any other dynamic classifier selection method.   
It is important to mention that the use of these techniques
not only improves the recognition rate but also can decrease the computational cost. Even for the methods
that have a fixed neighborhood size and therefore does not need to re-compute, the use of the algorithms
can still reduce the computational cost because the ENN eliminates some training patterns. Therefore it reduces the
cost of computing the nearest neighbor rule which can be high in some cases.




%

\bibliographystyle{IEEEbib}
\bibliography{ijcnn}

\end{document}